\documentclass[conference]{IEEEtran}

\renewcommand{\baselinestretch}{0.94}
\usepackage{cite}
\usepackage{amsmath,amssymb,amsfonts}
\usepackage{algorithmic}
\usepackage{graphicx}
\DeclareGraphicsExtensions{.pdf,.png}
\usepackage{textcomp}
\usepackage{xcolor}

\usepackage{soul}
\usepackage{color}
\usepackage{lipsum}
\usepackage{placeins}

\usepackage{textcomp}
\usepackage{rotating}
\usepackage{url}
\usepackage{xurl}
\usepackage{multirow}
\usepackage{gensymb}

\usepackage{booktabs}  % For professional tables
\usepackage{array}     % For column formatting

\usepackage{epstopdf}
\usepackage{graphicx}
\usepackage{booktabs}
\ifCLASSINFOpdf
\else
\fi
\begin{document}
\title{Predicting Train Delays in Finland \\Using Machine Learning and Weather Data\\}
%\title{Machine Learning for Train Delay Prediction \\by Integrating Weather Data: \\A Case Study of Railways in Finland\\}

\author{\IEEEauthorblockN{Vinicius Pozzobon Borin, Jean Michel de Souza Sant'Ana and Nurul Huda Mahmood}
\IEEEauthorblockA{Centre for Wireless Communications, University of Oulu, Oulu, Finland\\
\{vinicius.borin, jean.desouzasantana, nurulhuda.mahmood\}@oulu.fi}
}


%Telephone: (800) 555--1212, Fax: (888) 555--1212}


\maketitle

\begin{abstract}
Reliable railway operations depend increasingly on real-time environmental intelligence delivered through wireless sensor infrastructures, a capability that 6G networks will substantially enhance through integrated sensing and edge computing. Adverse weather, particularly in Arctic regions with extreme temperatures and heavy precipitation, remains a leading cause of train delays, yet most prediction approaches rely on raw meteorological inputs without exploiting domain-informed feature engineering. This paper investigates machine learning for train delay prediction using the Finland Integrated Train-Weather (FI-TW) dataset, which fuses railway operational records with observations from the Finnish Meteorological Institute's nationwide sensor network of approximately 200 stations communicating over wireless links. We evaluate three feature configurations using XGBoost at Oulu central station ($101{,}146$ observations): full weather features, instant weather observations only, and derived weather category scenarios. The category-based approach, employing hierarchical classifications such as Blizzard, Heavy Snow, and Extreme Cold, achieved an $\mathrm{R}^2$ of $0.78$, root mean squared error of $8.5$~minutes, and mean absolute error of $3.7$~minutes, representing an $11\%$ $\mathrm{R}^2$ improvement and $10\%$ error reduction over alternative configurations. These results demonstrate that compact, domain-informed features derived from sensors streams outperform raw meteorological observations, offering bandwidth-efficient representations suitable for edge deployment over current and emerging wireless infrastructures.
\end{abstract}

\begin{IEEEkeywords}
6G applications, logistics, data science, machine learning, rail transportation.  
\end{IEEEkeywords}


% , also known as the coefficient of determination, is a statistical measure used in machine learning and regression analysis to evaluate the goodness of fit of a model

% \fi
\IEEEpeerreviewmaketitle

\section{Introduction}

The evolution toward sixth-generation (6G) wireless networks is expected to enable data-driven services beyond traditional communications, including intelligent transportation and autonomous cyber-physical systems~\cite{Gkoumas2023}. By offering massive connectivity, ultra-reliable low-latency communication, and native support for sensing and artificial intelligence, 6G will underpin large-scale environmental monitoring infrastructures that feed operational decision-making with real-time data~\cite{Wang2023_6G}. Railway transportation, serving billions of passengers annually, stands to benefit significantly, as dense Internet-of-Things sensor networks along transportation corridors can anticipate disruptions and optimize operations.

Despite rail's growing popularity due to its sustainability and cost-effectiveness, train delays remain a significant challenge, causing passenger inconvenience, financial losses, and network congestion. These disruptions arise from technical malfunctions, operational constraints, infrastructure degradation, and environmental factors~\cite{Mukunzi2024, Monsuur2021}. Adverse weather conditions are particularly challenging due to their unpredictability and impact on service reliability, especially in Arctic regions. Consequently, accurate delay prediction becomes a critical operational tool, enabling proactive resource allocation, dynamic rescheduling, and improved maintenance planning.

The complexity of factors contributing to train delays makes conventional model-based approaches insufficient~\cite{Mukunzi2024}. Machine learning (ML) models trained on operational and environmental data can identify complex patterns and enable proactive interventions~\cite{Davari2021}. Despite extensive railway datasets covering traffic planning, maintenance, and safety~\cite{Pappaterra2021}, most do not fuse weather information with operational records. Huang et al.~\cite{Huang2021} proposed FCLL-Net for Chinese high-speed rail by incorporating raw meteorological observations as direct inputs, showing that excluding weather features increased Mean Absolute Error (MAE) by $2$--$4\%$, confirming the measurable contribution of meteorological data.

\begin{figure}[!bp]
    \centering
    \includegraphics[width=\columnwidth]{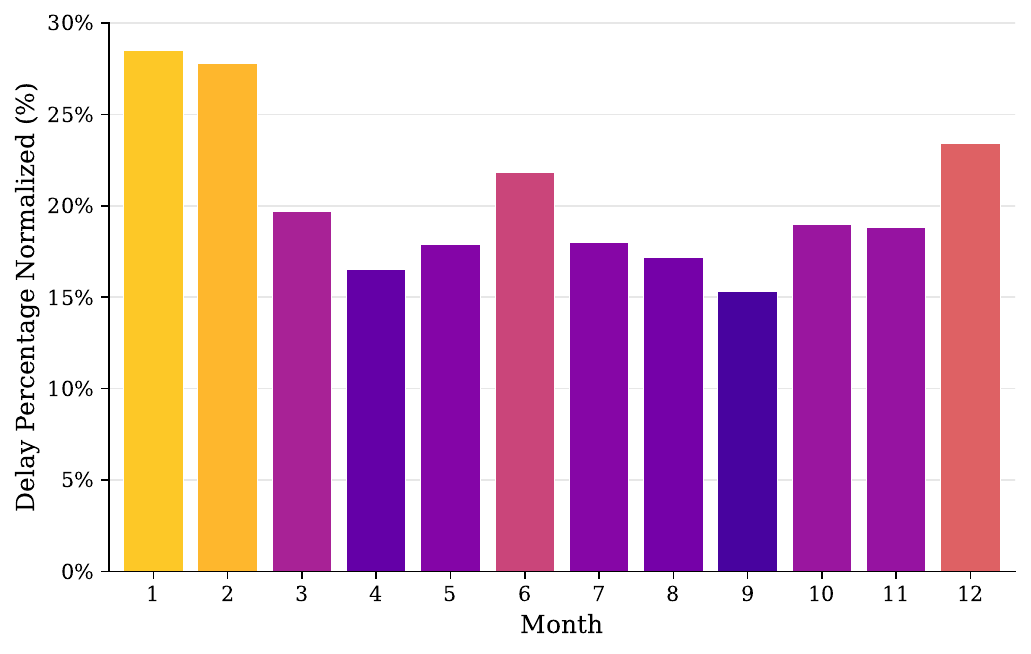}
    \caption{Monthly distribution of train delays (2018--2024). Values represent normalized delay percentage for each month across the study period.}
    \label{fig:delays_month}
    \vspace{-0.2cm}
\end{figure}

Within Europe, Finland's railway network presents a compelling case for weather-impact studies. The $5{,}915$-kilometer network serves over $90$ million passengers annually~\cite{ftia2024railway} and experiences extreme conditions, with winter temperatures as low as $-40$°C causing mechanical failures, signal disruptions, and reduced adhesion~\cite{Lotfi2023}. Environmental data is collected through the Finnish Meteorological Institute's (FMI) distributed sensor infrastructure, a nationwide IoT network of approximately 200 stations communicating over cellular wireless links, many in remote arctic locations where 5G/6G connectivity is essential for reliable data backhaul. Although $86.28\%$ of long-distance trains met punctuality standards in 2024~\cite{vayla2025railway}, monthly analysis reveals pronounced temporal variation. Figure~\ref{fig:delays_month} shows peak delay rates during winter months (Dec--Feb), with January reaching $28.5\%$ of normalized delay occurrences versus an autumn average of $17.7\%$.

To address this gap, this paper presents an ML approach for train delay prediction using the Finland Integrated Train-Weather (FI-TW) dataset (2018--2024)~\cite{fitw_dataset}, which fuses railway operational records with environmental observations from the FMI's nationwide sensor network, whose stations rely on cellular wireless connectivity for continuous data delivery. This work demonstrates how data collected and transmitted over current and future wireless infrastructure, including emerging 5G/6G networks, can be transformed into actionable intelligence for transportation systems. Specifically, our contributions include: (1) evaluation of ML architectures on weather-integrated railway data, (2) analysis of meteorological features' predictive value for delay forecasting, (3) a spatial-temporal fusion methodology for integrating heterogeneous sensor streams transmitted over wireless links with operational records, and (4) a hierarchical weather categorization system that transforms raw sensor measurements into bandwidth-efficient, operationally relevant features suitable for edge deployment in extreme arctic conditions.

\section{Dataset Description}

The FI-TW dataset~\cite{fitw_dataset} integrates railway operational records from Finland's Digitraffic railway traffic service with meteorological observations from FMI. Spanning January 2018 to December 2024, the dataset encompasses about $38.5$ million records distributed across Finland's railway network.

\subsection{Data Sources}

\subsubsection{Railway Operational Data}
The railway operational data provides comprehensive real-time and historical information about trains operating throughout Finland. The dataset includes train timetables with scheduled and actual departure/arrival times and station infrastructure metadata for Finland's railway network. The geographical distribution comprises $549$ stations within Finnish territory, of which $38.1\%$ serve passenger traffic while $61.9\%$ are designated for freight handling, docks, or technical service points. The functional classification reveals a hierarchical structure with conventional stations constituting $81.07\%$, stopping points representing $11.43\%$, and turnouts in the open line accounting for $7.50\%$. For each train journey, the dataset records both arrival and departure events at each station along the route. This distinction allows for precise tracking of dwell time at stations and delay propagation throughout the journey. A journey with $n$ stopping points generates $2n-1$ rows when all stations involve commercial stops, with the origin station containing only a departure record and the final destination containing only an arrival record. The delay at each observation point is calculated as $\textit{differenceInMinutes} = \textit{actualTime} - \textit{scheduledTime},$ 
% \begin{equation}
% \textit{differenceInMinutes} = \textit{actualTime} - \textit{scheduledTime},
% \end{equation}
where positive values indicate delays, zero indicates on-time performance, and negative values indicate ahead-of-schedule arrivals.

\subsubsection{Meteorological Data}
Meteorological data is sourced from $209$ FMI environmental monitoring stations distributed across Finland. These stations are configured at two distinct measurement intervals: $164$ stations ($78.47\%$) operate with 10-minute intervals following World Meteorological Organization standards~\cite{WMO-No82008}, while $45$ stations ($21.53\%$) provide high-resolution 1-minute measurements typically positioned at locations requiring detailed temporal resolution such as aviation meteorology~\cite{SKYbrary2024WeatherObservations}. The integration of weather and train data involves a two-stage process. First, spatial matching is performed using the Haversine formula~\cite{sinnott1984virtues}, which computes great-circle distances between each train segment's geographic coordinates and all available meteorological stations operated by the FMI. Each train segment is then assigned to the nearest station, minimizing spatial interpolation error and ensuring that the meteorological observations reflect local conditions as accurately as possible. Second, temporal alignment synchronises the operational and meteorological records by matching each train observation to the closest available weather measurement in time. Since FMI stations report observations at fixed intervals, this step resolves any offset between the train event timestamp and the meteorological recording window, producing a fully aligned dataset in which every observation is associated with both the geographically nearest and temporally closest weather record.

\subsection{Dataset Features}
The final dataset comprises $39$ features organized into three categories: target ($5$), operational ($13$), and weather ($21$) features. Table~\ref{tab:dataset_features} provides a comprehensive overview of all features, where the ``Derived'' column indicates features that were engineered from existing raw features through mathematical transformations or aggregations.

\begin{table}[!bp]
\centering
\caption{Dataset Features Overview}
\label{tab:dataset_features}
%\resizebox{\textwidth}{!}{%
\begin{tabular}{clll}
\hline
\textbf{No.} & \textbf{Feature Name} & \textbf{Type} & \textbf{Derived} \\
\hline
1 & differenceInMinutes & Target & \\
2 & differenceInMinutes\_offset & Target & X \\
3 & differenceInMinutes\_eachStation\_offset & Target & X \\
4 & trainDelayed & Target & X \\
5 & cancelled & Target & \\
6 & trainStopping & Operational & \\
7 & commercialStop & Operational & \\
8 & month & Operational & \\
9 & month\_sin & Operational & X \\
10 & month\_cos & Operational & X \\
11 & hour\_sin & Operational & X \\
12 & hour\_cos & Operational & X \\
13 & hour & Operational & \\
14 & day\_of\_week & Operational & \\
15 & day\_week\_sin & Operational & X \\
16 & day\_week\_cos & Operational & X \\
17 & day\_of\_month & Operational & \\
18 & train\_id & Operational & \\
19 & Air temperature & Weather & \\
20 & Wind speed & Weather & \\
21 & Gust speed & Weather & \\
22 & Wind direction & Weather & \\
23 & Relative humidity & Weather & \\
24 & Dew-point temperature & Weather & \\
25 & Precipitation intensity & Weather & \\
26 & Snow depth & Weather & \\
27 & Pressure at mean sea level & Weather & \\
28 & Horizontal visibility & Weather & \\
29 & Cloud amount & Weather & \\
30 & weather\_scenario\_Normal\_Clear & Weather & X \\
31 & weather\_scenario\_Blizzard & Weather & X \\
32 & weather\_scenario\_Heavy\_Snow & Weather & X \\
33 & weather\_scenario\_Extreme\_Cold & Weather & X \\
34 & weather\_scenario\_Heavy\_Rain & Weather & X \\
35 & weather\_scenario\_Freezing\_Rain & Weather & X \\
36 & weather\_scenario\_Black\_Ice & Weather & X \\
37 & weather\_scenario\_Dense\_Fog & Weather & X \\
38 & weather\_scenario\_High\_Winds & Weather & X \\
39 & weather\_scenario\_Extreme\_Heat & Weather & X \\
\hline
\end{tabular}%
%}
\end{table}

\subsubsection{Target features}
The dataset includes five target variables designed to capture different aspects of railway operational performance. The primary target variable \textit{differenceInMinutes} represents the cumulative delay in minutes from the start of the journey until the current station where the data is collected, capturing the total accumulated delay. The \textit{differenceInMinutes\_offset} variant removes the initial delay recorded at the first station, isolating delays accumulated during the journey. The \textit{differenceInMinutes\_eachStation\_offset} further removes delay propagation effects at each intermediate station, focusing only on the delay between two adjacent stations. Two binary target variables complement these numeric targets: \textit{trainDelayed} indicates whether a train experienced any delay, while \textit{cancelled} identifies cancelled train services.

\subsubsection{Operational features} capture temporal patterns and train service characteristics. Temporal variables (\textit{hour}, \textit{month}, \textit{day\_of\_week}) are encoded using sine-cosine transformations to preserve cyclical relationships, addressing the time wraparound problem where values at cycle boundaries appear maximally distant despite temporal proximity~\cite{Cai2020}. The encoding follows
\begin{equation}
t\text{\_sin} = \sin\left(\frac{2\pi t}{P}\right)~\text{and}~ t\text{\_cos}=\cos\left(\frac{2\pi t}{P}\right),
\end{equation}
where $t$ is the temporal value and $P$ is the period (24 for hours, 12 for months, 7 for days).

\subsubsection{Weather Features}
The weather feature set comprises 11 meteorological measurements obtained from FMI stations: air temperature (°C), wind speed (m/s), gust speed (m/s), wind direction (degrees), relative humidity (\%), dew-point temperature (°C), precipitation intensity (mm/h), snow depth (cm), pressure at mean sea level (hPa), horizontal visibility (m), and cloud amount (oktas). These measurements provide continuous numerical values representing atmospheric conditions at the time of train operations. Given the absence of globally standardized definitions for severe weather phenomena such as blizzards, heavy snow, or freezing rain (or even consensus at the European level), we developed a hierarchical classification system to categorize operationally relevant weather conditions. Our approach draws inspiration from European transport weather impact research~\cite{vajda2014severe}, while representing a novel application of categorical weather classification for railway delay prediction in extreme climatic conditions. We derived 10 additional binary features from meteorological measurements through hierarchical classification logic that categorizes weather conditions into mutually exclusive scenarios based on combined thresholds of multiple parameters. The classification follows a severity-based priority order to ensure proper categorization when conditions overlap, as detailed in Table~\ref{tab:weather_categories}.

\begin{table}[!bp]
\centering
\caption{Weather Category Classification Criteria}
\label{tab:weather_categories}
\small
\begin{tabular}{@{}lp{5.5cm}@{}}
\toprule
\textbf{Category} & \textbf{Classification Criteria} \\
\midrule
\textit{Blizzard} & Temp. $<0$°C, Precip. $>1$ mm/h or $>3$ mm, wind $>10$ m/s or gust $>15$ m/s, visibility $<1000$ m \\
\textit{Heavy Snow} & Temp. $<0$°C, Precip. $>2$ mm/h or $>5$ mm, snow depth $>0$ cm \\
\textit{Extreme Cold} & Temp. $<-20$°C \\
\textit{Heavy Rain} & Temp. $>2$°C, Precip. $>4$ mm/h or $>10$ mm \\
\textit{Freezing Rain} &  $-2\degree \text{C} < \text{temp.} < 2\degree \text{C}$, Precip. $>0$ mm/h or $>0$ mm\\
\textit{Black Ice} & $-2\degree \text{C} < \text{temp.} < 2\degree \text{C}$, humidity $>80$\%, dew-point - temp $<2\degree\text{C}$, precip. $>0$ \\
\textit{Dense Fog} & Precip. $\leq0.1$ mm, Visibility $<1000$ m, humidity $>95$\% \\
\textit{High Winds} & Wind $>15$ m/s or gust $>20$ m/s \\
\textit{Extreme Heat} & Temp. $>30$°C \\
\textit{Normal/Clear} & No severe weather conditions met \\
\bottomrule
\end{tabular}
\end{table}

This hierarchical approach ensures that the most operationally disruptive conditions (e.g., blizzards) take precedence in classification, preventing misclassification when multiple threshold criteria are simultaneously met.

\section{Machine Learning Training Setup}

This section presents the XGBoost model selection and experimental setup, followed by correlation-based feature analysis that informs the definition of three training scenarios with different feature configurations.

\subsection{Model Selection: XGBoost}

This study employs XGBoost (eXtreme Gradient Boosting)~\cite{chen2016xgboost} as the primary ML algorithm for delay prediction. XGBoost is an optimized implementation of gradient boosting that constructs an ensemble of decision trees sequentially, where each subsequent tree corrects the residual errors of the preceding ensemble. The algorithm has demonstrated favorable performance across diverse tabular data applications and is particularly well-suited for datasets combining numerical and categorical features with complex non-linear relationships.

Several characteristics make XGBoost appropriate for railway delay prediction. The algorithm inherently handles mixed feature types and missing values, reducing preprocessing requirements. Its tree-based structure naturally captures non-linear interactions between weather conditions and operational variables without explicit feature engineering. Additionally, XGBoost provides built-in feature importance metrics that enable interpretation of which meteorological and operational factors most strongly influence delay predictions, supporting both predictive accuracy and domain understanding.

\subsection{Experimental Setup}

The experiments focus on Oulu asema (Oulu central station), identified in the network analysis as a high-delay node ($19.0\%$ delay rate). The target variable is \textit{differenceInMinutes} (Feature~1 in Table~\ref{tab:dataset_features}), representing the cumulative delay in minutes per train event. The dataset comprises all long-distance services arriving or departing at Oulu asema during 2018--2024,
totalling $101{,}146$ observations, which capture the seasonal and meteorological variability of northern Finland.

Because railway operations and weather both exhibit strong temporal autocorrelation, a uniformly random partition would leak future information into training. We therefore apply a chronological $80/20$ split: the earliest $80\%$ of observations (sorted by event timestamp) form the development set, and the most recent $20\%$ are held out as a
final test set, used only once for reporting. Within the development set, model selection uses 5-fold expanding-window time-series cross-validation, where each validation fold strictly succeeds its training fold in time. Robust scaling and all other preprocessing transformations are fitted on each training fold in isolation and then
applied to the subsequent validation fold, so no statistic derived from future data influences fitting. Hyperparameter tuning employs randomized search over the space in Table~\ref{tab:hyperparameters},
drawing $100$ candidate configurations; the configuration minimising the mean validation RMSE across the time-series folds is retrained on the full development set and evaluated once on the held-out test set.

Table~\ref{tab:hyperparameters} presents the hyperparameter search space. The number of estimators was sampled uniformly between $100$ and $400$, while maximum tree depth ranged from 4 to 8 to balance model complexity and generalization. Learning rates of $0.01$, $0.05$, and $0.1$ were evaluated alongside subsampling ratios for both observations and features. The \textit{scale\_pos\_weight} parameter was tuned to address class imbalance in the binary delay classification task, with values reflecting the approximate ratio of non-delayed to delayed observations.

\begin{table}[!bp]
\centering
\caption{XGBoost Hyperparameter Search Space}
\label{tab:hyperparameters}
\begin{tabular}{@{}ll@{}}
\toprule
\textbf{Parameter} & \textbf{Distribution/Values} \\
\midrule
n\_estimators & Uniform integer [100, 400] \\
max\_depth & Uniform integer [4, 8] \\
learning\_rate & \{0.01, 0.05, 0.1\} \\
subsample & \{0.7, 0.8, 0.9\} \\
colsample\_bytree & \{0.7, 0.8, 1.0\} \\
scale\_pos\_weight & \{3.9, 4.9, 5.9\} \\
\bottomrule
\end{tabular}
\end{table}

The base operational feature set, used across all experimental scenarios, comprises 8 features that capture fundamental scheduling and temporal characteristics independent of weather conditions. This set includes train operational indicators (\textit{trainStopping} and \textit{train\_id}) and cyclical temporal encodings (sine and cosine components for month, hour, and day of week). The complete specifications for these features are provided in Table~\ref{tab:dataset_features} (Features 6, 9--12, 15--16, 18).

\subsection{Feature Analysis and Selection}

\begin{figure}[!bp]
    \centering
    \includegraphics[width=\columnwidth]{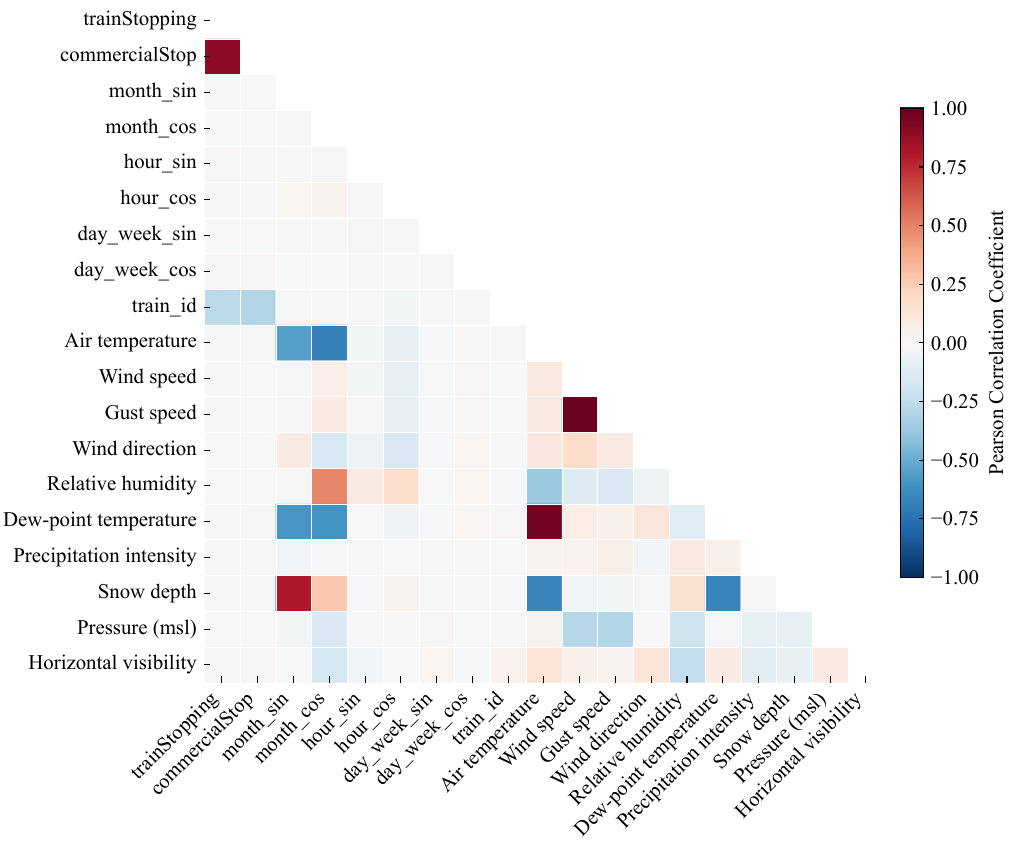}
    \caption{Correlation heatmap of operational and weather features in the training dataset.}
    \label{fig:correlation_heatmap}
\end{figure}

A comprehensive correlation analysis was conducted to identify potential feature redundancies and inform feature selection strategies. Figure~\ref{fig:correlation_heatmap} presents the correlation matrix for all features in the training dataset, excluding cloud amount which was dropped due to data unavailability at Oulu Asema. We also excluded the derived weather scenarios from this analysis. In the heatmap, highly correlated features are represented by dark red (positive correlation) or dark blue (negative correlation), while similar lighter colors indicate weaker positive or negative correlations, respectively.

The analysis reveals several notable patterns with implications for model design. Among weather variables, gust speed and wind speed exhibit very strong correlation ($r = 0.977$), indicating near-redundancy between these features. Similarly, dew-point temperature shows extremely high correlation with air temperature ($r = 0.966$), reflecting the thermodynamic relationship between these variables. 

Seasonal patterns emerge clearly in the temperature-related correlations. Air temperature correlates negatively with month encoding components (month\_sin: $r = -0.549$; month\_cos: $r = -0.678$), capturing Finland's pronounced seasonal temperature variation. Snow depth exhibits strong positive correlation with month\_sin ($r = 0.805$), and strong negative correlation with air temperature ($r = -0.658$) and dew-point temperature ($r = 0.663$), reflecting winter accumulation patterns. Relative humidity shows moderate positive correlation with month\_cos ($r = 0.490$) and negative correlation with air temperature ($r = -0.369$).

Atmospheric pressure demonstrates moderate negative correlations with wind speed ($r = -0.287$) and gust speed ($r = -0.295$), aligning with meteorological principles where low-pressure systems are associated with stronger winds. Horizontal visibility correlates negatively with relative humidity ($r = -0.246$), reflecting reduced visibility during humid conditions such as fog or precipitation events.

Notably, operational features (\textit{trainStopping}, cyclical time encodings, \textit{train\_id}) show weak correlations with weather variables, with most coefficients below $|0.1|$. This independence suggests that weather features provide complementary information not captured by operational variables alone, supporting the hypothesis that meteorological data can enhance delay prediction beyond what temporal and operational patterns provide.

\subsection{Training Scenarios}
Based on the correlation analysis presented in Figure~\ref{fig:correlation_heatmap}, feature selection was performed to reduce redundancy while preserving predictive information. Highly correlated  variables were excluded. The \textit{commercialStop} feature was removed due to its redundancy with \textit{trainStopping}. Additionally, \textit{gust speed} and \textit{dew-point temperature} were excluded from all training scenarios due to their high correlations with wind speed and air temperature, respectively, to mitigate multicollinearity effects.

Based on this analysis, three feature configuration strategies were evaluated:
\begin{enumerate}
    \item \textit{Full weather features set}: Comprehensive feature set combining nine instant meteorological observations with 10 weather category scenarios, alongside operational features.
    \item \textit{Instant weather observations only}: Operational features combined exclusively with nine instant meteorological observations only.
    \item \textit{Weather category scenarios only}: Operational features paired solely with the 10 derived weather scenario categories listed in Table~\ref{tab:weather_categories}.
\end{enumerate}

Table~\ref{tab:training_scenarios} summarizes the feature composition for each training scenario. All configurations utilize nine operational features, comprising train stopping status, cyclical temporal encodings, day of month, and train identifier.

\begin{table}[!bp]
\centering
\caption{Feature Composition Across Training Scenarios}
\label{tab:training_scenarios}
\footnotesize
\setlength{\tabcolsep}{3pt}
\begin{tabular}{p{1.5cm}p{1.6cm}p{1.6cm}p{1.6cm}}
\toprule
\textbf{Feature Category} & \textbf{Scenario 1 \newline Full Set} & \textbf{Scenario 2 \newline Instant Only} & \textbf{Scenario 3 \newline Categories Only} \\
\midrule
Operational Features      & 6, 9--12, \newline 15--16, 18 & 6, 9--12, \newline 15--16, 18 & 6, 9--12, \newline 15--16, 18 \\
\midrule
Instant Weather Observations & 19--20, \newline 22--23, 25--29 & 19--20, \newline 22--23, 25--29 & -- \\
\midrule
Weather Category Scenarios & 30--39 & -- & 30--39 \\
\midrule
\textbf{Total Features}   & \textbf{28} & \textbf{18} & \textbf{19} \\
\bottomrule
\end{tabular}
\end{table}

\section{Training Results and Discussion}

Because our target variable is numeric (train delay in minutes), the metrics employed in this analysis are defined as follows:
\begin{itemize}
    \item \textit{R² (Coefficient of Determination):} Measures the proportion of variance in the target variable explained by the model, with values closer to $1.0$ indicating superior explanatory power.
    
    \item \textit{RMSE (Root Mean Square Error):} Quantifies the standard deviation of prediction residuals, penalizing larger errors more heavily due to the squared term. Expressed in minutes.
    
    \item \textit{MAE (Mean Absolute Error):} Represents the average magnitude of prediction errors without considering their direction, providing a linear and interpretable measure in the same units as the target variable (minutes).
    
    \item \textit{WMAPE (Weighted Mean Absolute Percentage Error):} Expresses the total absolute error as a percentage of total actual values, offering a scale-independent assessment particularly useful when comparing across datasets of different magnitudes. Unlike the traditional MAPE, which divides by individual actual values and becomes undefined or unreliable when delays approach zero, WMAPE aggregates all actual values in the denominator. This distinction is critical in railway delay prediction, where trains frequently operate on schedule (zero delay) or experience minimal delays. WMAPE avoids the instability and asymmetry inherent in MAPE by preventing division by near-zero values.
\end{itemize}

\begin{figure*}[!htbp]
    \centering
    \includegraphics[width=0.75\textwidth]{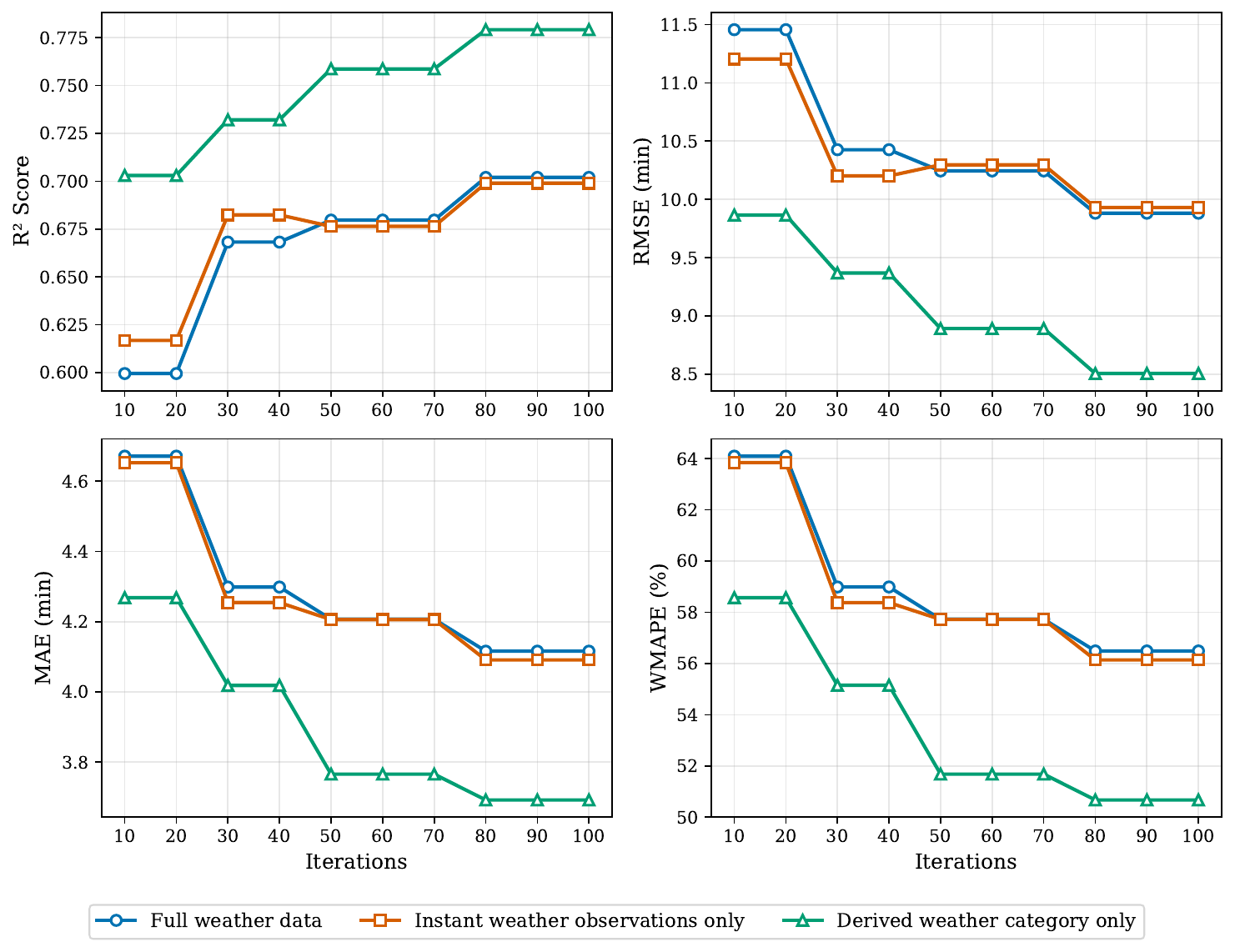}
    \caption{Performance metrics (R², RMSE, MAE, WMAPE) across training iterations for the three feature scenarios.}
    \label{fig:performance-iterations}
\end{figure*}

Figure~\ref{fig:performance-iterations} presents the evolution of these four metrics across training iterations for each feature scenario. The results demonstrate consistent performance improvements across all scenarios as the number of iterations increases. The which incorporates only the meaningful weather categories, consistently outperforms the other two scenarios across all metrics. At $100$ iterations, this scenario achieves an R\textsuperscript{2} of approximately $0.78$, compared to roughly $0.70$ for other two scenarios. This represents a meaningful improvement in explanatory power, suggesting that engineered weather category features contribute more to delay prediction accuracy than raw weather observations alone.

Examining the error metrics reveals similar patterns. The RMSE for Scenario~3 decreases from approximately $9.9$ minutes at $10$ iterations to $8.5$ minutes at $100$ iterations, whereas Scenarios~1 and~2 converge to approximately $9.9$ minutes. The MAE follows a comparable trajectory, with Scenario~3 reaching approximately $3.7$ minutes compared to $4.1$ minutes for the baseline scenarios. The WMAPE metric shows that Scenario~3 achieves roughly $50.5\%$ at convergence, representing a notable improvement over the $56.5\%$ observed for Scenarios~1 and~2.

A notable observation is the convergence behavior of the models. All scenarios exhibit rapid improvement during the initial iterations, with the most substantial gains occurring between 20 and 50 iterations. Scenarios 1 and 2 plateau by approximately 50 iterations, beyond which additional training yields negligible improvement. Scenario 3 continues to improve modestly until approximately 80 iterations, after which performance stabilizes across all metrics. Since doubling the iteration count from 50 to 100 yields only marginal additional gains, the 50--80 iteration range represents a practical stopping point that balances computational cost against predictive performance. Furthermore, Scenarios 1 and 2 exhibit nearly identical performance throughout the iteration range, indicating that the features distinguishing these scenarios contribute minimally to predictive accuracy compared to the weather variables introduced in Scenario 3.

\subsection{Comparison with Related Work}

Direct numerical comparison with Huang et al.~\cite{Huang2021} requires acknowledging key methodological differences. Their evaluation partitioned delays into short ($4-30$ minutes) and long ($>30$ minutes) categories and reported metrics per partition, whereas our approach trains and evaluates on the full continuous distribution of delay values. Their routes also exhibit considerably lower delay variability, with average delays of $3.12$ and $1.10$ minutes respectively, while Finland's $5,915$~km network experiences substantially higher variability, with the proportion of delayed trains reaching approximately $27.5\%$ in winter compared to $18\%$ in summer, a near twofold seasonal difference driven largely by severe Nordic weather conditions.

Our Scenario~3, XGBoost model achieved MAE of $3.7$ minutes, RMSE of $8.5$ minutes, and R\textsuperscript{2} of $0.78$. While this MAE is higher than the $0.659$ minutes reported by Huang et al.~\cite{Huang2021} for short delays, their evaluation targeted a narrower and less variable delay range on high-speed routes, whereas our model operates on the full continuous delay distribution of a network experiencing severe winter conditions. Given these substantially more challenging operating conditions, the results demonstrate competitive performance and confirm that engineered weather category, rather than raw meteorological observations, constitute effective inputs for ML-based delay prediction across distinct railway contexts and climatic conditions.

\subsection{Generalization Beyond Oulu Station and Limitations}

Although evaluation is restricted to Oulu central station, the proposed weather categorization is designed to generalize. The thresholds reflect physical mechanisms of railway disruption, such as rail adhesion loss near the freezing band, catenary stress under high winds, and signalling impairment under low visibility, which are invariant to geographic location. The same definitions therefore remain operationally meaningful in milder climates, where category frequencies simply shift toward rain, fog, or black-ice conditions. Binary indicators are also more robust to domain shift than raw continuous observations, since the operational regime carries a consistent signal regardless of the underlying meteorological distribution. The framework draws inspiration from pan-European transport weather research~\cite{vajda2014severe}, supporting its applicability across heterogeneous IoT sensor deployments.

This study has four explicit limitations. (i) Empirical validation across additional stations and railway networks remains future work. (ii) Only XGBoost is benchmarked; alternative ML and DL architectures are not assessed. (iii) Each train segment is matched to its nearest FMI station via the Haversine distance, ignoring micro-climatic variability and multi-station interpolation. (iv) Reliable wireless backhaul from FMI sensors is assumed, with missing observations treated as data gaps rather than communication failures, an assumption that may not hold in low-coverage arctic deployments.

\section{Conclusion}

This paper presented baseline ML experiments for train delay prediction using the FI-TW dataset, evaluating three feature configuration strategies with XGBoost models at Oulu central station.
The experimental results demonstrate that employing derived weather category scenarios rather than raw meteorological observations consistently outperforms both the full feature set and instant weather observations only across all evaluation metrics. At convergence, this achieved an R\textsuperscript{2} of $0.78$, RMSE of $8.5$ minutes, and MAE of $3.7$ minutes, representing approximately $14\%$ improvement in RMSE and $10\%$ improvement in MAE compared to the alternative configurations.
Beyond predictive performance, training the model on derived weather category offers practical advantages for deployment. The weather category features are encoded as binary indicators rather than continuous floating-point values, substantially reducing memory requirements and storage footprint. This efficiency becomes particularly relevant for large-scale railway networks where millions of observations must be processed and stored.
Building on the limitations identified above, future research should extend the evaluation to additional stations, benchmark DL architectures such as recurrent neural networks and transformers against the XGBoost baseline, and jointly model meteorological inputs with the reliability of their wireless backhaul.


\section*{Acknowledgment}

{\small This work was supported by European Union through the Interreg Aurora project ENSURE-6G (Grant Number: $20361812$) and the Research Council of Finland (former Academy of Finland) through the 6G Flagship program (Grant Number: $369116$)}



%\bibliographystyle{IEEEtran}
%\bstctlcite{IEEEexample:BSTcontrol}
\bibliographystyle{IEEEtran.bst}
\bibliography{IEEEabrv,references}

\end{document}